\documentclass[conference]{IEEEtran}

\usepackage{cite}
\usepackage{comment}
\usepackage{amsmath,amssymb,amsfonts}
\usepackage{algorithmic}
\usepackage{graphicx}
\usepackage{textcomp}
\usepackage{xcolor}
\usepackage{booktabs}
\usepackage{cleveref}
\usepackage{url}

\newcommand{\etal}{\textit{et al.}~}

\newcommand{\eg}{\textit{e.g.},~}

\def\BibTeX{{\rm B\kern-.05em{\sc i\kern-.025em b}\kern-.08em
    T\kern-.1667em\lower.7ex\hbox{E}\kern-.125emX}}
\begin{document}

\title{AutoSIGHT: Automatic Eye Tracking-based System for Immediate Grading of Human experTise}

\author{\IEEEauthorblockN{Byron Dowling}
\IEEEauthorblockA{\textit{University of Notre Dame} \\
Notre Dame, IN, USA \\
bdowlin2@nd.edu}
\and
\IEEEauthorblockN{Jozef Porubcin}
\IEEEauthorblockA{\textit{University of Notre Dame} \\
Notre Dame, IN, USA \\
jporubci@nd.edu}
\and
\IEEEauthorblockN{Adam Czajka}
\IEEEauthorblockA{\textit{University of Notre Dame} \\
Notre Dame, IN, USA \\
aczajka@nd.edu}
}

\maketitle

\begin{abstract}
Can we teach machines to assess the expertise of humans solving visual tasks automatically based on eye tracking features? This paper proposes AutoSIGHT, Automatic System for Immediate Grading of Human experTise, that classifies expert and non-expert performers, and builds upon an ensemble of features extracted from eye tracking data while the performers were solving a visual task. Results on the task of iris Presentation Attack Detection (PAD) used for this study show that with a small evaluation window of just 5 seconds, AutoSIGHT achieves an average average Area Under the ROC curve performance of 0.751 in subject-disjoint train-test regime, indicating that such detection is viable. Furthermore, when a larger evaluation window of up to 30 seconds is available, the Area Under the ROC curve (AUROC) increases to 0.8306, indicating the model is effectively leveraging more information at a cost of slightly delayed decisions. This work opens new areas of research on how to incorporate the automatic weighing of human and machine expertise into human-AI pairing setups, which need to react dynamically to nonstationary expertise distribution between the human and AI players (\eg when the experts need to be replaced, or the task at hand changes rapidly). Along with this paper, we offer the eye tracking data used in this study collected from 6 experts and 53 non-experts solving iris PAD visual task.
\end{abstract}



\section{Introduction}
As Artificial Intelligence (AI) systems become more commonplace in everyday tasks, companies and researchers alike understand that a lack of trust in a model or the validity of a model's decision is a major obstacle to wide-scale adoption \cite{hasija2022artificial}. This has led to the sub-field of Trustworthy Artificial Intelligence (TAI) that focuses on defining the core principles that AI systems should satisfy to increase trust and adoption. One such principle is that good models should generalize well to unseen data types (that is, operate well in an open set recognition regime). Another principle is that there should exist a seamless and effective collaboration between the AI and humans solving the tasks jointly, in which the capabilities of both sides are appropriately and automatically assessed, and incorporated into the decision-making process.

One family of techniques addressing the first principle and aiming at increasing generalization of AI models solving visual recognition tasks, in which humans are competent, is to utilize human saliency (a map of features supporting human's decisions), to improve fidelity of model's feature maps \cite{Dang_2020_CVPR}, or to directly guide the model's learning process to increase model's generalization capabilities \cite{Boyd_2023_WACV}. However, human saliency-based training methods often include non-expert participants, with assuming that general visual capabilities offered by humans may provide sufficient guidance in the saliency-based training \cite{crum2023explain, boyd2022human}. Additionally, although some studies have attempted to embed expert eye gaze into AI models \cite{Sultana_2024_ACCV, kaushal2023detecting}, these studies are limited to medical domains where a high level of expertise of participating subjects is presumed. However, the methodologies of these studies cannot apply to domains where there is a lack of consensus on who the expert performers are. 

This paper focuses on the second principle and offers a method of automatic expertise assessment of human subjects solving an example biometric presentation attack detection task, that is, judging whether an iris image represents a real eye or a fake object. A particular application of this method is a real-time weighing of the AI's and human's expertise level in a scenario where (a) AI and humans collaborate to deliver a decision about a visual sample they process, and (b) the expertise level of either side may change rapidly (for instance, when the human expert needs to be replaced during the task, or the task at hand changes). 

This brings up an important question, what characteristics determine if someone is an expert in a domain? For medical domains such as radiology, expertise is defined by obtaining a medical doctorate and several years of experience viewing radiologic images \cite{samei2018handbook, krupinski2013characterizing, kelly2016development}. However, in other domains such as biometric presentation attack detection or anomaly detection, the differentiating characteristics of who is an expert are less clear. Additionally, a robust method for distinguishing the differences in visual expertise \textit{between} experts is lacking. The ability to discern between expert and non-expert performers is the catalyst to empowering several important applications in the VL/HCC areas that include, but are not limited to: smarter source code review policies to prevent project disruption, educational tools that dynamically adjust content difficulty to match a student's current performance, and Human-Centric AI tools that can continuously monitor a participant's performance and identify fatigue. Therefore, it is necessary to design a classifier and methodology that can properly and in real time assess expert performers in visual inspection tasks. 

Thus, in this paper, we present AutoSIGHT, an Automatic System for Immediate Grading of Human experTise. The main contributions of this work are: 
\begin{enumerate}
    \item the first known to us expert/non-expert paradigm experiment for iris presentation attack detection task;
    \item a classification architecture and methodology for assessing visual expertise directly from eye tracking data, with a potential application to weigh human experts and AI methods in human-machine pairing settings;
    \item eye tracking dataset that contains raw gaze data along with gaze heatmaps and fixations in an expert/non-expert paradigm (in which both experts and non-experts solved the same visual task) experiment.
\end{enumerate}

This paper is organized around the following research questions:

\begin{itemize}
    \item \textbf{(RQ1)} What characteristics distinguish expert and non-expert visual performers in the domain of iris presentation attack detection?
    \item \textbf{(RQ2)} Can we train a classifier to automatically and in real-time distinguish experts and non-experts using eye tracking data from visual inspection tasks?
    \item \textbf{(RQ3)} How long does a participant need to be observed by the system for an accurate expertise rating?
\end{itemize}

We offer the source codes, trained models and data along with this paper for reproducibility purposes and to facilitate follow-up research efforts\footnote{\url{https://github.com/CVRL/AutoSIGHT}}.

%
\section{Related Work}
The motivation for the development of the AutoSIGHT methodology is to use eye tracking to assess if someone is displaying visual expertise during a particular sequence. If the participant is adjudged to be an expert on that particular sequence, then that sequence can be leveraged as expert human saliency to inform future models in expert-saliency-guided training. Therefore, it is important to understand the current state of works in the trustworthy AI area, how human saliency is being used to address explainability and generalization concerns, and what eye tracking can tell us about the differences in visual processing of information between experts and non-experts. 

\subsection{Trustworthy AI}
Social interaction involves some degree of trust \cite{hobbes2016leviathan}. In the past few years, different sections of society have embraced AI and have started to integrate it into critical tasks. Despite wide-spread success in many areas such as Natural Language Processing, Medicine, and Biometrics, this success has raised new concerns about the trustworthiness surrounding AI decisions and the processes involved. In fact, one can argue that the majority of AI relies on implicit trust rather than explicit trust. These concerns surrounding the degree of trust have resulted in the new sub-field of trustworthy AI with several guidelines and strategies for actively building trust in AI systems  \cite{thiebes2021trustworthy, smuha2019eu, brundage2020trustworthy, 10.1145/3491209, kaur2021requirements, li2023trustworthy, liu2022trustworthy}. On the principle of explainability, one particular problem that has manifested from wide-scale AI adoption is the problem of modern models being ``black boxes'',  and thus offering limited interpretability of AI decisions. This issue comes in two different forms, the first being whether a user will trust an individual prediction of a model, and the second being whether the user will trust a model to behave in reasonable ways \cite{ribeiro2016should}. The issue is further exacerbated depending on what tasks are given for AI to assist with. Consumers have shown tendency to trust AI for tasks that appear to be objective in nature such as providing directions or predicting the weather, however with more subjective tasks such as diagnosing a disease, a person is likely to trust a human expert over AI \cite{castelo2019task}. Resistance to AI in medical use is particularly prevalent as various studies have shown higher levels of trust in expert human doctors over AI use in medicine \cite{longoni2019resistance, longoni2020resistance, robertson2023diverse, riedl2024patients}. Therefore, despite the recent breakthroughs in the performance of black box AI, the increases in classification performance has not lead to an equivalent increase in trust in the prediction and performance of these models \cite{jiang2018trust, nguyen2015deep}.

\subsection{Model and Human Saliency}
A popular technique for increasing explainability in the decision process of a Deep Neural Network (DNN) is to apply Gradient-weighted Class Activation Mapping (Grad-CAM) to the final convolutional layer. Grad-CAM is a method that creates a coarse localization map that highlights the important regions within an image that ultimately led to the model's decision. The final output is a heatmap over the original image that shows what pixels within the image factored most into the decision \cite{selvaraju2017grad}. Grad-CAM is not without flaws however. One weakness is that it will occasionally highlight features or locations within images that were not actually used in the model decision \cite{draelos2021use}. These visualizations are commonly used by human examiners to determine whether the model is following conventional feature detection. Models that show high classification accuracy and CAMs with irrelevant features are more likely to have focused on accidental information in training and may struggle to generalize on unseen data \cite{crum2023explain}.

One technique that aims to increase generalization of the AI models, and build upon the usefulness of Grad-CAM, is to utilize human saliency, an attention map with regions of interest, to improve model feature maps \cite{Dang_2020_CVPR}, or to directly guide the learning process to make model's and human's saliencies pointing to the similar features \cite{Boyd_2023_WACV}. Despite the success of Grad-CAM, when benchmarking different saliency methods for chest X-ray interpretation, while Grad-CAM localized pathologies performed better than the other tested saliency methods, all seven of the tested saliency methods, including Grad-CAM, performed quite poorly against a human expert benchmark \cite{saporta2022benchmarking}. However, early attempts to blend model saliency and human saliency in the field of biometrics have shown promising results \cite{Boyd_2023_WACV}. Human saliency was captured via hand-marked annotations which was used to steer model saliency to focus more on areas deemed salient by humans. The increased classification performance while arriving at decisions from human guidance adds a level of trust seldom seen in AI models \cite{boyd2023value}. 

Despite this success, by using human annotators of varying skill and expertise, the blended saliency maps force the model to look at erroneous features that an expert may have disregarded within their own annotation. Additionally, the participants that provided the annotations were given instructions to annotate five regions that support their decisions. These instructions make sense from a standpoint that they guarantee you will receive substantial annotations. On the other hand, forcing the annotators to select five regions per image likely disproportionately gives credence to the importance of these regions in the identification process. It is possible, if not probable, that there were instances where a human annotator felt that only two regions contributed to their decision on the image, but had to arbitrarily select three additional ones to comply with the instructions. Likewise, there could have been a scenario where the annotator wanted to highlight more than five regions but was constrained to only five. With these limitations, eye tracking appears to be a more natural solution that accurately captures relevant features and decision making processes.

\subsection{Eye Tracking and Human Expertise}
Expertise has many definitions, but the consensus is that it refers to high levels of performance on a particular task or within a particular domain \cite{bourne2014expertise}. A popular technique for evaluating visual expertise involves using eye tracking devices to record and capture data to be post-processed and interpreted. Eye tracking has been employed in human perception research to better identify gaze and eye movement during visual tasks \cite{jarodzka2021eye, borys2017eye}, as well as numerous studies comparing experts and non-experts in specific domains \cite{tien2014eye,gegenfurtner2011expertise, kosel2023measuring, castner2020pupil, bednarik2018pupil}. However few studies include experts whose qualifications are not backed by medical degrees. 

Instead, many studies compare the performance of experienced medical professionals with several years of practicing medicine against ones who are still in residency \cite{li2023using, brunye2019eye, castner2020pupil, bednarik2018pupil, bernal2014experts, capogna2020novice}. These studies between the expert and novice groups conclude with subtle differences in the average fixation duration, fixation duration and count within Areas of Interest (AOIs), and changes in pupil size. These findings follow the current state of literature in eye tracking for visual expertise in which domain experts tend to log more fixations and more fixations of shorter duration than their novice counterparts \cite{gegenfurtner2011expertise, ooms2014study}, which distinguishes expert and non-expert performance in motor tasks \cite{kosel2023measuring, gegenfurtner2011expertise, ooms2014study, dogusoy2014cognitive, li2023using}. 

However, there are gaps in the literature towards the real-time assessment of visual expertise. To the best of our knowledge, there are no studies that extract \textit{features} known to be good predictors of visual expertise and utilize them to train AI models to automatically classify expertise. There are very few, if any, studies that seek to predict what category of visual expertise an individual participant demonstrates in visual inspection tasks and whether that performance fluctuates between or throughout tasks. Finally, there are very few, if any, studies that seek to assess the differences \textit{between} experts and determine which showed better visual expertise. With the AutoSIGHT system, we aim to identify eye tracking features that are good predictors of visual expertise and use these features to train an AI model to make expertise assessment automatically, and in doing so present a methodology that can be used to potentially differentiate the performance between between experts.

\subsection{Detection of Biometric Presentation Attacks}
Iris recognition is widely regarded as one of the most secure biometric methods and continues to be deployed in large-scale applications including personal identification, border control, among others \cite{boyd2020iris, czajka2018presentation, daugman2009iris}. As with any secure system, there are always attempts to undermine the system's integrity, and iris recognition is no exception \cite{czajka2018presentation}. These attempts are often referred to as presentation attacks which involves presenting a spoofed sample to the biometric sensor with the intent of manipulating the system into an incorrect decision \cite{czajka2018presentation}. The intent behind these attacks ordinarily comes in two forms: identity concealment, where the attacker wishes to evade detection, or impersonation, where the goal is to spoof the biometric system with a positive match under false pretenses \cite{boyd2023comprehensive}. 

There are many different methods of presentation attacks to iris biometric systems including presenting high resolution printouts of authentic irises to the scanner \cite{pacut2006aliveness}, producing a post-mortem or cadaver sample\cite{trokielewicz2018presentation}, or synthetic iris imagery using Generative Adversarial Networks (GANs) such as RaSGAN, StyleGAN2, and StyleGAN3 \cite{Yadav_2019_CVPR_Workshops, karras2020analyzing, karras2021alias}. With the variety of these sophisticated spoofing techniques comes a need for countermeasures in the form of \textit{presentation attack detection}. Deep learning-based approaches have experienced success at the detection of presentation attacks by learning representations directly from data \cite{menotti2015deep}, deep feature extraction \cite{minaee2016experimental, alaslani2018convolutional}, fine tuning existing models trained for general-purpose vision problems \cite{boyd2019deep}, and more recently, using human-aided saliency maps to improve generalization to unseen data \cite{boyd2022human}. Despite the success of these methods, they are still black box in nature and have the tendency to pick up on irrelevant features \cite{kim2019saliency}. The goal of AutoSIGHT is to classify the visual expertise of the participant, utilize the eye tracking gaze data to determine expert-human-saliency, to improve presentation attack detection while restraining the model from picking up on erroneous features.

                                                                                
                                                               

\section{Data Collection Design}

\subsection{Overview}

\begin{figure*}[ht]
    \centerline{\includegraphics[width=\textwidth]{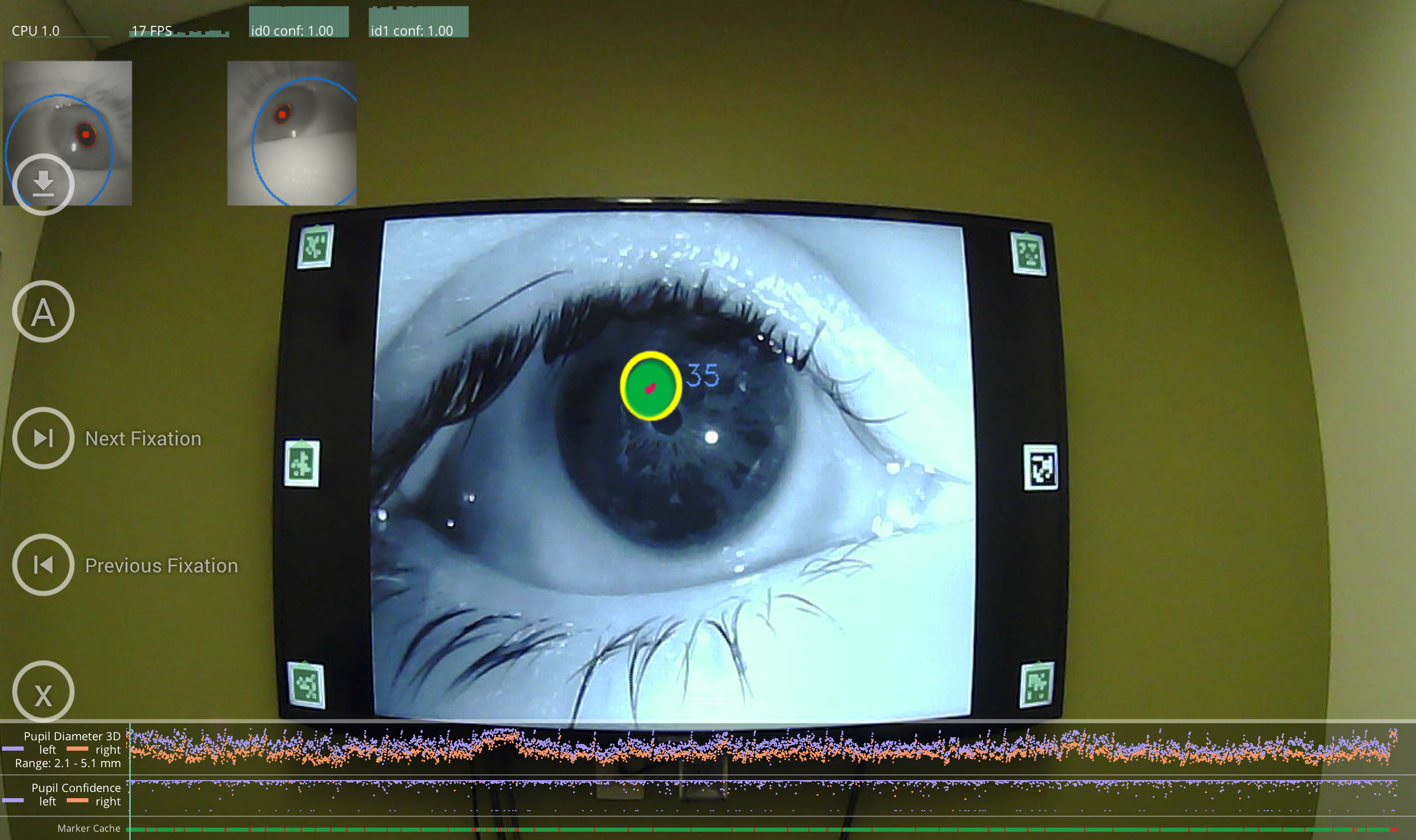}}
    \caption{Example frame from the PupilCore recording of the experiment in progress with a participant attempting to correctly identify a sample as ``Abnormal'' (in this case iris showing signs of pupil miosis).}
    \label{fig:session}
\end{figure*}

Our study population consisted of 8 expert and 70 non-expert participants solving the visual task of detecting anomalies in biometric iris images (related to attacks, such as printed irises or covered by textured contact lenses, and related to biology-sourced anomalies, such as eye diseases or post-mortem changes). Age, gender, ethnicity and race were recorded for all participants. The expert participants were recruited from the university faculty doing an active research in iris presentation attach detection, as well as from the ophthalmology research community. The criteria for this initial expert moniker were if the participants possessed one or more years of specialized iris recognition-related research, or possessed a medical doctorate in ophthalmology. The 70 non-expert participants consisted of university students recruited from various majors as well as university staff, all of whom did not possess any specialized biometrics research experience. This study was approved by the Institutional Review Board (IRB), and each participant signed a consent form allowing us to release anonymized eye tracking-sourced data.

\begin{figure*}[ht]

    \centering
    \begin{minipage}{0.23\textwidth}
        \includegraphics[width=\linewidth]{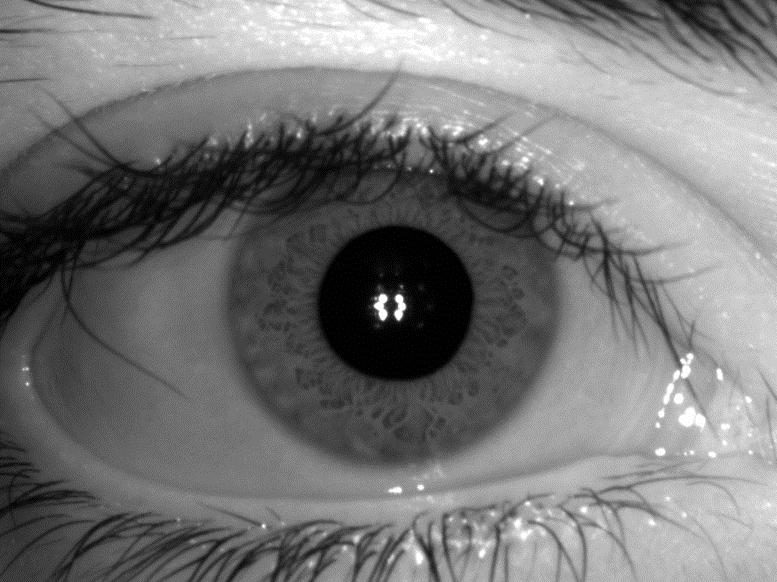}
    \end{minipage} \hspace{0.01\textwidth}
    \begin{minipage}{0.23\textwidth}
        \includegraphics[width=\linewidth]{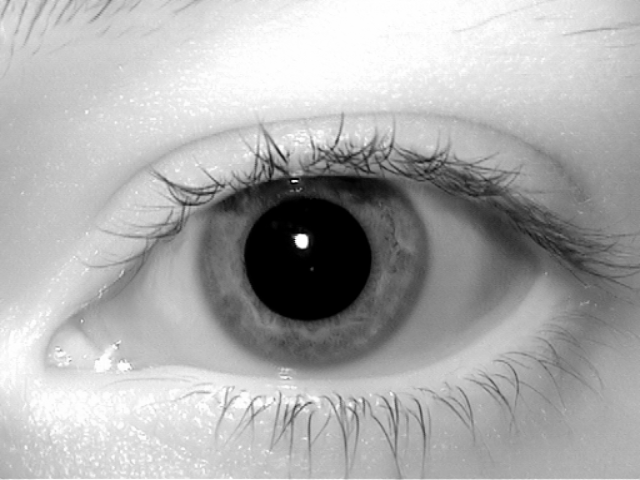}
    \end{minipage} \hspace{0.01\textwidth}
    \begin{minipage}{0.23\textwidth}
        \includegraphics[width=\linewidth]{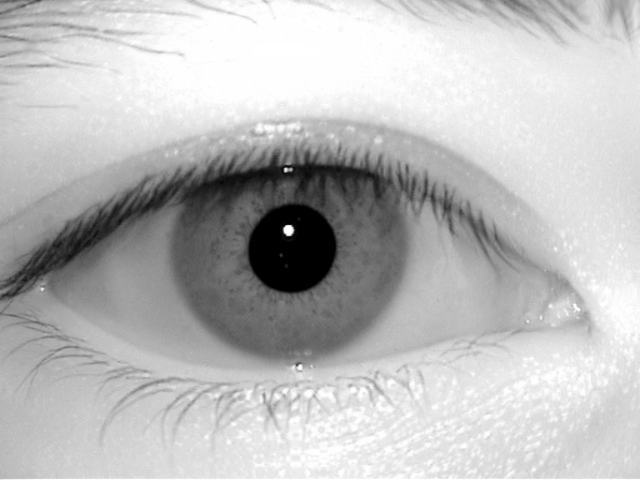}
    \end{minipage} \hspace{0.01\textwidth}
    \begin{minipage}{0.23\textwidth}
        \includegraphics[width=\linewidth]{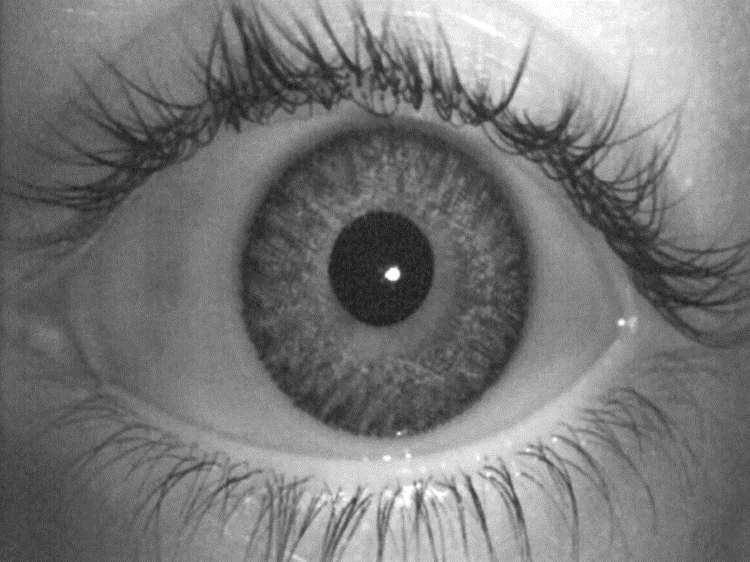}
    \end{minipage}

    \vspace{0.2cm} 

    \begin{minipage}{0.23\textwidth}
        \includegraphics[width=\linewidth]{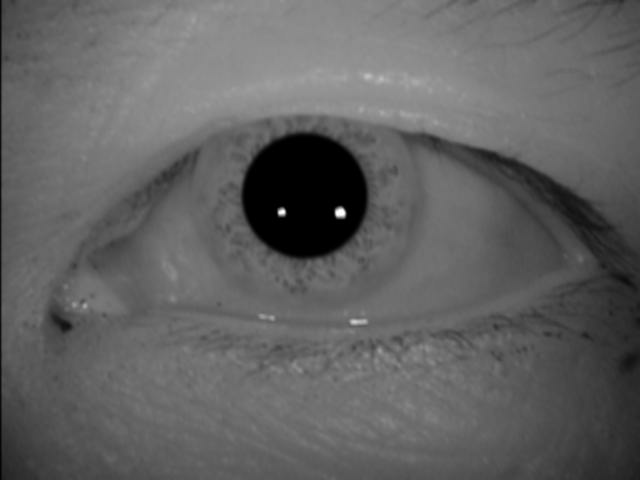}
    \end{minipage} \hspace{0.01\textwidth}
    \begin{minipage}{0.23\textwidth}
        \includegraphics[width=\linewidth]{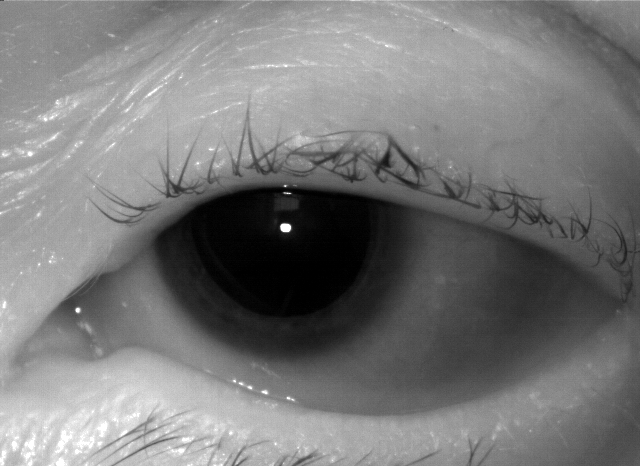}
    \end{minipage} \hspace{0.01\textwidth}
    \begin{minipage}{0.23\textwidth}
        \includegraphics[width=\linewidth]{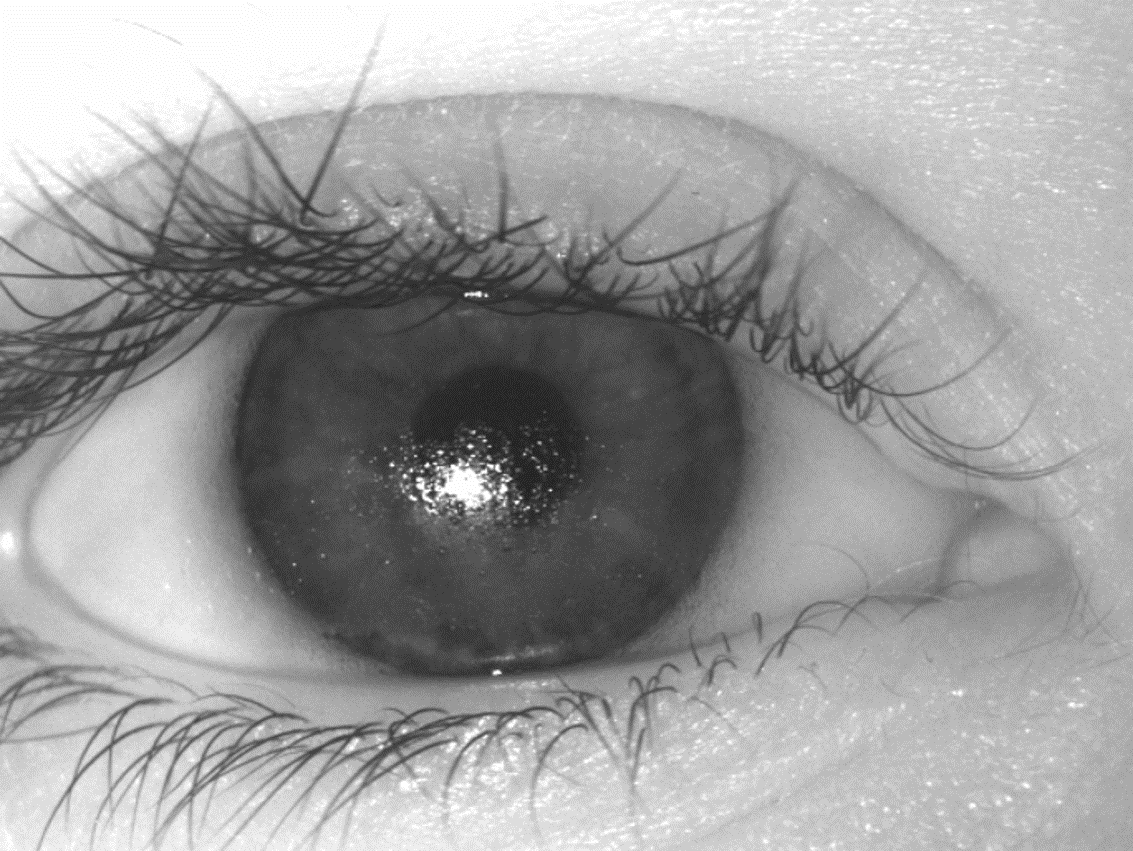}
    \end{minipage} \hspace{0.01\textwidth}
    \begin{minipage}{0.23\textwidth}
        \includegraphics[width=\linewidth]{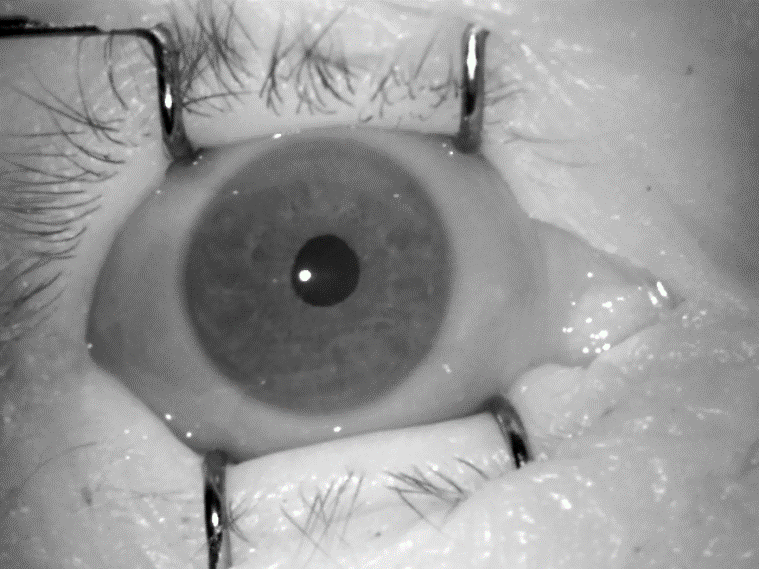}
    \end{minipage}

    \vspace{0.2cm} 

    \begin{minipage}{0.23\textwidth}
        \includegraphics[width=\linewidth]{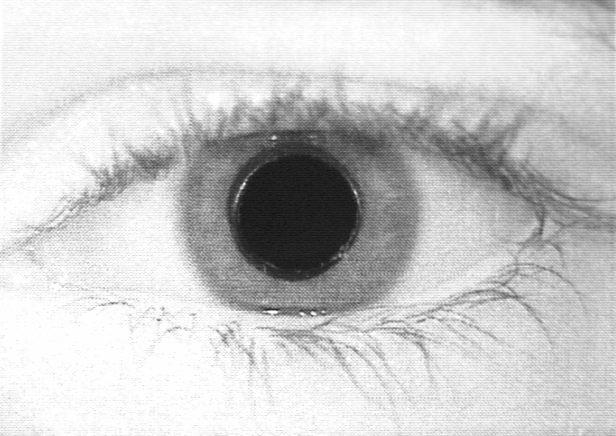}
    \end{minipage} \hspace{0.01\textwidth}
    \begin{minipage}{0.23\textwidth}
        \includegraphics[width=\linewidth]{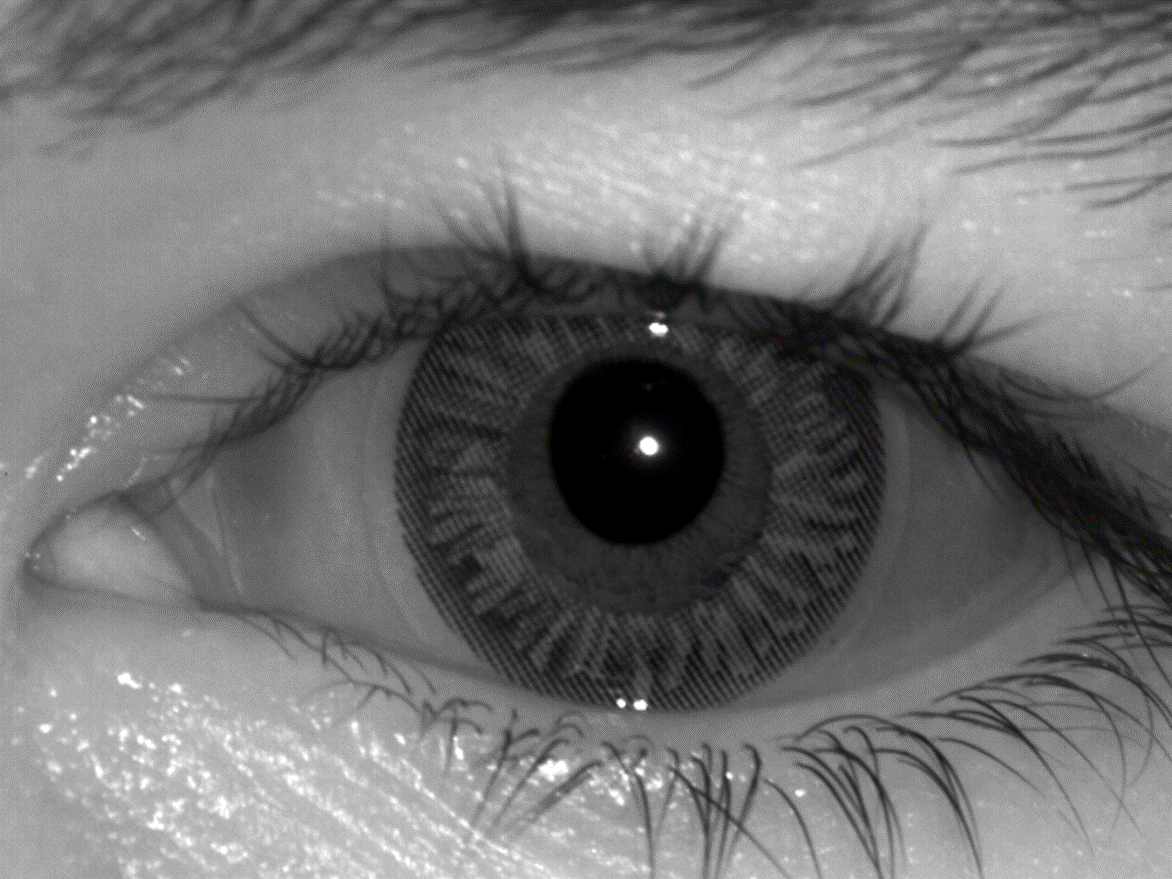}
    \end{minipage} \hspace{0.01\textwidth}
    \begin{minipage}{0.23\textwidth}
        \includegraphics[width=\linewidth]{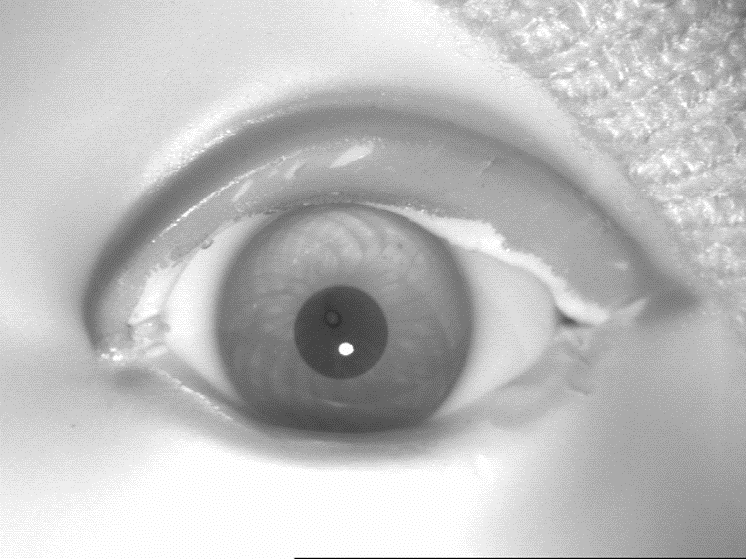}
    \end{minipage}

    \caption{From left to right (with third-party dataset sources, where appropriate): live iris (with no abnormalities) \cite{Kohli_BTAS_2016}, StyleGAN2-generated sample, StyleGAN3-generated sample, iris wearing textured contact lens, then printed and re-captured in near infrared light \cite{Kohli_BTAS_2016}, synthetic sample generated by a non deep learning-based algorithm \cite{CASIA_Synth}, diseased eye \cite{Trokielewicz_BTAS_2015}, glass prosthesis, post-mortem sample \cite{Trokielewicz_TIFS_2019}, iris printout \cite{Czajka_MMAR_2013}, iris wearing textured contact lens \cite{Doyle_IPBR_2014}, and artificial eye \cite{Kim_ESA_2016}.}
    \label{fig:attackTypes}
\end{figure*}

\subsection{Minimal Reporting Guidelines for Eye Tracking Research}
In adherence to the minimal reporting guidelines for eye tracking research and to maximize reproducibility of our methodology, we outline the eye tracking-specific reporting guidelines in this section \cite{dunn2024minimal}. All eye tracking sessions were recorded using a PupilLabs PupilCore headset tethered to a Windows laptop running the latest version of {\it Pupil Capture} software to record and save the session recordings. PupilCore samples at 200 Hz for both left and right eyes by default, however some participant recordings were limited to one eye video during capture to ensure the capture of reliable and high quality data only. All participant sessions took place in the same conference room where participants sat in a chair approximately 1.5 meters away from a 55-inch TV monitor, far enough away so that the entire screen could be viewed without moving the head, and was positioned to minimize environment lighting affects as demonstrated in Fig. \ref{fig:session}. Additionally, participants were given strict instruction not to move their heads once they became comfortably situated in the chair for the duration of the experiment. Calibration was conducted before the start of each experiment for each participant by utilizing the built-in calibration functionality of {\it Pupil Capture}. Upon completion, each session was post-processed utilizing PupilLab's {\it Pupil Player} software and its ``3D Gaze'' estimation pipeline. Parameters recorded include raw gaze positions and fixation data of which other parameters were extrapolated from including but not limited to fixation durations, counts, and normalized $(x,y)$ position locations. 

\subsection{Distribution of Images}

The images used in the experiment are sourced from the dataset used previously in a human annotation study conducted by Boyd \etal \cite{boyd2022human}. The images in this set, in addition to bonafide iris images, represent the following attack types and iris image anomalies: diseased eyes, irises with textured contact lenses, artificial (plastic) eyes, professional glass prosthesis, printouts, irises wearing textured contact lenses then printed and re-captured by an iris sensor, synthetically-generated iris images, and post-mortem iris images, Fig.~\ref{fig:attackTypes}. Additionally, 100 synthetic iris images were generated, 50 from StyleGAN2 and 50 from StyleGAN3 models \cite{tinsley2023iris, karras2020analyzing, karras2021alias}, and added to the set bringing the final amount to 864 images. Of these attack types, only bonafide live irises are considered ``Normal'' in the context of our study, all other anomalies represent the ``Abnormal'' category. These images were divided into 16 unique decks comprised of 54 unique images each. Each deck was balanced by anomaly type to prevent any one deck from appearing too easy or too difficult. Each deck contained on average 6 diseased iris images, 5 textured contact lens images, 2 glass prosthesis images, 1 artificial iris image, 6 printout images, 5 contact lens and printed images, 5 synthetic (non-StyleGAN) images, 4 post-mortem iris images, 12 bonafide images, 3 StyleGAN2-generated images, and 3 StyleGAN3-generated images. During preliminary test sessions prior to the official collection, StyleGAN2, bonafide irises, and StyleGAN3 were the three most challenging categories to answer correctly.

\subsection{Data Acquisition Procedure}
Each session of the collection process began with a calibration of the PupilCore headset for accurate measurements, optimal microphone placement for accurate recording of responses, and accommodations for maximum comfort during the collection process. Upon completion of this phase, all subjects regardless of expertise level, were given a brief training prior to the start of the experiment that showed an example image of all the attack types present in the dataset and clarity on what constitutes ``Normal'' and ``Abnormal'' categories. These training samples were excluded from the actual experiment. Subjects were then given the opportunity to ask any clarifying questions regarding the experiment, which were answered within reason to not give any unfair advantages to the participants. 

During the actual experiment, each participant was shown 54 unique irises from the deck and was asked to examine each image using all features contained within the image, and verbally announce whether they believed the image to be ``Normal'' or ``Abnormal.'' Upon completion of this initial phase, participants were then asked to verbally describe what led to them making that decision during the initial phase. Once completed, participants were instructed to announce their final decision. The moderator would advance to the next sample in the deck. The procedure did not have a time limit. Participants were encouraged to take as much time as needed to examine the images during the initial phase and to be as brief or thorough as they felt necessary to describe how they arrived at their answers. This was done primarily to ensure participants did not examine or identify features out of compliance with an arbitrary time frame or anomaly count as described in prior human annotation studies \cite{boyd2023value}, but instead to authentically capture the examination process. On average, a single session lasted approximately 30 minutes. 

\subsection{Collected Data}

\begin{table*}[ht]
\centering
\caption{Overview of Dataset Features}
\label{tbl:dataset_features} 
\begin{tabular}{p{3cm}|p{2.5cm}|p{3cm}|p{4cm}}
    \toprule
    \textbf{Feature Name} & \textbf{Type} & \textbf{Range/Units} & \textbf{Description} \\
    \midrule
    Participant ID & Identifier & N/A & Unique pseudo-identifier randomly assigned to each subject, guarantying participants' anonymity \\
    \hline
    Image Source and Ground Truth Label & Identifier & N/A & Image source and label of what attack type category it belongs to \\
    \hline
    Expertise Level & Categorical & Expert / Non-Expert & Participant's expertise designation \\
    \hline
    Initial Decision & Categorical & ``Normal'' or ``Abnormal'' & Participant's response for initial decision \\
    \hline
    Final Decision & Categorical & ``Normal'' or ``Abnormal'' & Participant's response for final decision \\
    \hline
    Raw Gaze Data Pointer & Numerical & 1 - N & Sequence number that corresponds to a \verb+csv+ file containing raw gaze $(x,y)$ coordinates of the specified time window \\
    \hline
    Average Fixation Time & Numerical & 80 - 4000 ms & Average time participant's gaze remained on a single fixation point \\
    \hline
    Fixation Count & Numerical & 1 - N counts & Total number of fixations recorded during the task where N is the max total logged per image \\
    \hline
    Fixation Times & Numerical & 80 - 4000 ms & List of individual fixation durations logged for this sample \\
    \hline
    Gaze Relational Index (GRI) & Numerical & 0 - 1 normalized  & Ratio of Average Fixation Duration to fixation count on a per-image basis \\
    \hline
    Image Heatmap & Visual Output & 0-1 pixel intensity values & Heatmap of eye tracking gaze data overlayed on original image to highlight salient features \\
    \bottomrule
\end{tabular}
\end{table*}

At a high level, the dataset provides the following parameters for each image:

\begin{itemize}
    \item image source (reference to the original dataset and file name)
    \item ground truth label of the image anomaly type
    \item a participant's pseudo identifier and the subject's decision
    \item sequences of raw $(x,y)$ gaze coordinates in different sliding window sizes that occurred during evaluation of the image
    \item fixation times (for every fixation that occurred)
    \item fixation counts
    \item average fixation time
    \item Gaze Relational Index (GRI) score, which is the ratio of mean fixation time to total fixation count \cite{gegenfurtner2020gaze}.
    \item eye gaze heatmap
\end{itemize}

This data is structured as a list of JSON objects, where each file represents an individual participant, and each object in the list represents the raw statistics for that participant on a particular image. By compiling statistics and correlating data on a per-image basis, it allows for further research into how levels of expertise can vary throughout the experiment. The eye tracking gaze heatmaps can be used as human saliency maps in expert/non-expert paradigm experiments. A more streamlined summarization can be found in Table \ref{tbl:dataset_features}.

\subsection{Data Curation}
The data collection process started with 70 non-expert and 8 expert participants. Low-quality recordings (\eg from participants wearing glasses, which in certain circumstances made it impossible to properly calibrate the pupil core headset, or issues related to head shape, which caused the eye camera placement not supporting the participant's full range of visual scanning) were discarded, ending up with good-quality data collected from 53 non-expert and 6 expert participants.

\section{Experimental Design}

\subsection{Expertise Level Assessment Features}
\label{ELAF}
AutoSIGHT seeks to utilize eye tracking metrics that are generalizable, can be calculated in real time, and require minimal prior knowledge. While many traditional eye tracking metrics rely on AOIs such as Time To First Fixation, Dwell Time, or Revisits, they require predefined AOIs in order to be calculated. Therefore, these features are highly domain-specific and not conducive for rapid assessment of visual expertise. Likewise, although we have the ground-truth classification of each image the participant evaluated, we did not incorporate whether the participant correctly classified the image. Doing so would require any future applications using AutoSIGHT's methodology to not only require the end user to provide a classification of what they are visually inspecting, but also need an associated ground truth before an expertise rating could be generated. Such constraints would not be practical for many potential applications and thus we elected to exclude any ground truth features.

Pupil dilation has been cited as an indicator of expertise \cite{szulewski2017pupillometry, bednarik2018pupil, castner2020pupil}, however recent findings from Ruuskanen \etal \cite{ruuskanen2023baseline} found weak to non-existent relationship between variability in pupil size and expertise, both in average and variability of baseline, while controlling for several variables known to influence pupil size. Pupil size may also be affected by ambient lighting conditions, which may limit the number of application scenarios. The variability of these findings, combined with the need for rigorous control of external variables, both in data collection and in future applications, led to us not considering this feature. Instead, we have focused on eye tracking features that are more generalizable to any domain. These include Average Fixation Duration (AFD), computed average time of fixations during a sequence, Fixation Count (FC), the amount of fixations recorded in a sequence, raw gaze locations during evaluation which captures what features were used during the decision process, and average Euclidean distance (AED) between fixations which captures how saccade movements, fixation sequences, and search behavior play a role in the assessment of the image.

\subsection{Sliding Window Approach}
\label{sec:sliding_window}
We hypothesized in \textbf{RQ3} that there exists a minimum amount of time that a participant's viewing behavior needs to be observed before an accurate expertise prediction can be made. Therefore, to pre-process the data for these experiments, we took each participant's entire recording and segmented their eye tracking data into 5-, 10-, 15-, 20-, and 30-second windows utilizing a sliding window approach of advancing the window size by half. For example, a 10-second window size would start at the beginning of the experiment and segment relevant gaze and fixation data for this single sequence, then the window would advance 5 seconds so the next sequence would include 5 seconds of previous data and 5 seconds of new data. The data extracted for each sequence included the gaze coordinates, AFD, FC, and the AED between the fixations of that sequence. Sliding windows were generated until the end of the experiment was reached for this participant. The process was repeated for each participant for all the listed window sizes so that different observation window sizes could be tested.

\begin{figure}[!ht]
    \centerline{\includegraphics[width=\linewidth]{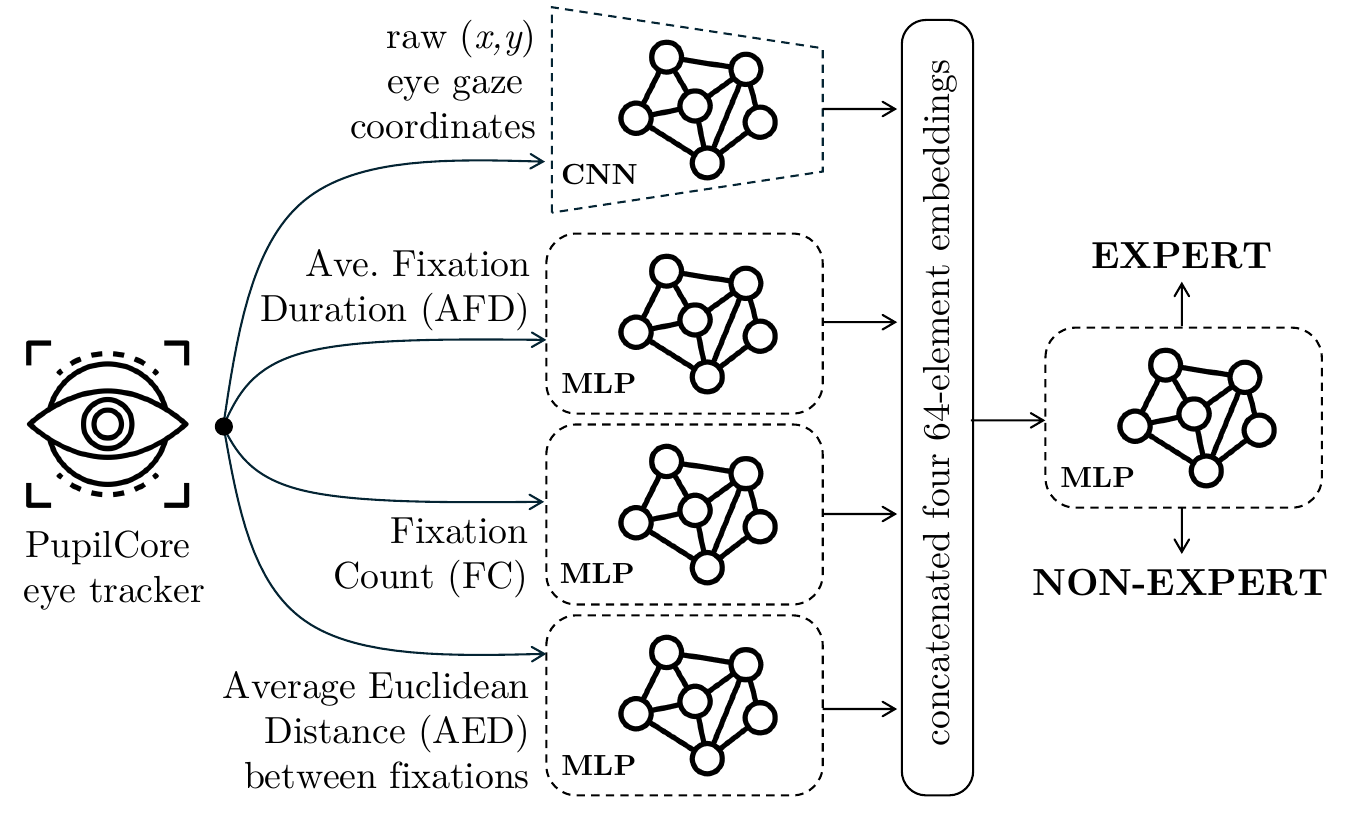}}
    \caption{The architecture of the proposed multi-stream neural network-based classifier, assessing the expertise of subjects solving the visual inspection task.}
    \label{fig:network}
\end{figure}

\subsection{Classifier Design}
We trained a binary classifier to predict the expertise level using a specially designed multi-stream neural network, as illustrated in Fig. \ref{fig:network}. A ResNet-based \cite{He_CVPR_2016} 1-dimensional CNN accepted a 2-channel sequence of normalized $(x,y)$ gaze coordinates with a fixed input size of the eye tracker's sampling rate multiplied by each tested window size. The AFD, FC, and AED metrics for a given time sequence of data were individually fed into separate multilayer perceptrons (MLP). The outputs of the three MLPs and the 1D CNN were concatenated into a final embedding vector, followed by an additional MLP making the final categorization into expert or non-expert classes based on the {\it softmax} score (1.0 denoting high expertise, while 0.0 denoting no expertise in a given domain). This estimation was done for each window within each sequence in a given session, what simulates a real-time (with the window size time granularity) estimation of the expertise level.

\subsection{Subject-Disjoint Train-Validation-Test Splits}
During the model training, to prevent overfitting to a specific participant's tendencies, we split the data to balance the number of experts and non-experts for training. We left out one expert and non-expert's session for validation, and additional one expert and non-expert's session for testing, while the data of the four remaining experts and four random non-experts were selected for training. This subject-disjoint train-validation-test regime is required to assess how this approach generalizes to unseen participant data, instead of learning properties of particular subjects.

\section{Results}

\subsection{Answering \textbf{RQ1}: What characteristics distinguish expert and non-expert performers?}
The aforementioned expert level assessment features (Sec. \ref{ELAF}), were compiled into distributions of expert features and non-expert features to assess for statistically-significant differences. Distributions followed the typical left-sloped long-tailed distribution probability densities for AFD, FC, and the AED described in previous works \cite{rayner1998eye, negi2020fixation}. We conducted a non-parametric Mann-Whitney U test (assuming the significance level $\alpha=0.05$) and found statistically significant differences between the distributions of features extracted for experts and non-experts for all three of these features with experts showing significantly lower AFD, a tendency to log more fixations than non-experts, and lower overall AED. {\bf Thus, answering RQ1, our results show that AFD, FC, and AED are statistically significant eye tracking-sourced features that correlate with expertise}. 

\subsection{Answering \textbf{RQ2}: Can we train a classifier to automatically and in real-time distinguish experts and non-experts using eye tracking data from visual inspection tasks?}
\begin{figure*}[ht]
    \centerline{\includegraphics[scale=0.53]{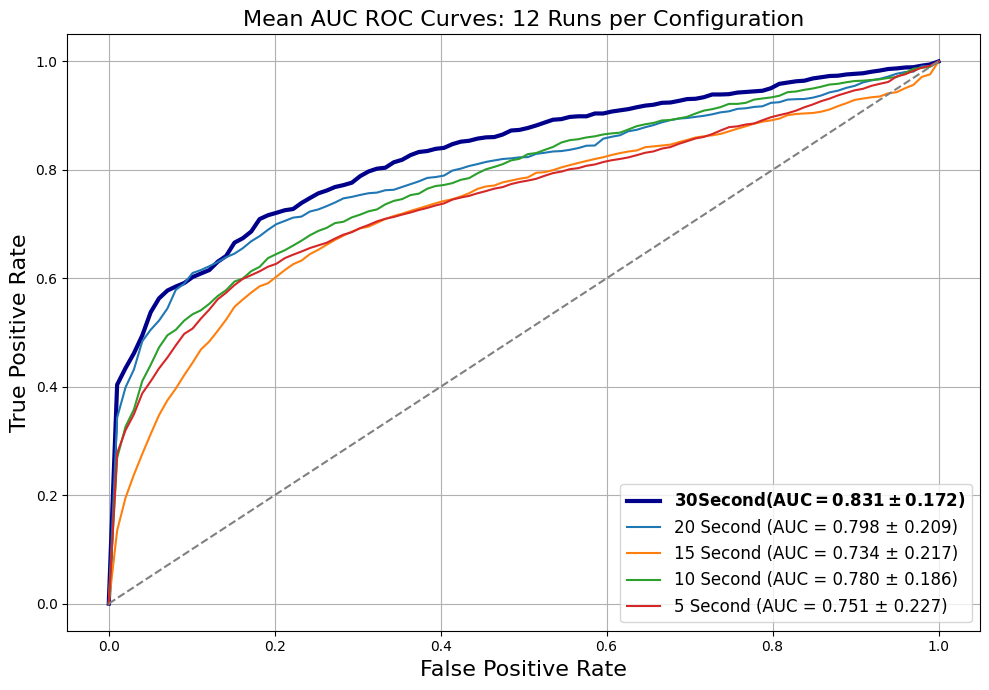}}
    \caption{ROC curves along with the average AUROC and standard deviation for the best performing configuration of each window size tested. Results outline that each configuration is able to separate expert and non-expert performers at a success rate that far exceeds random chance despite the relative ease of the task on many samples.}
    \label{fig:ROC_Curves}
\end{figure*}

\begin{figure*}[tp]
    \centering
    \includegraphics[width=0.9\linewidth]{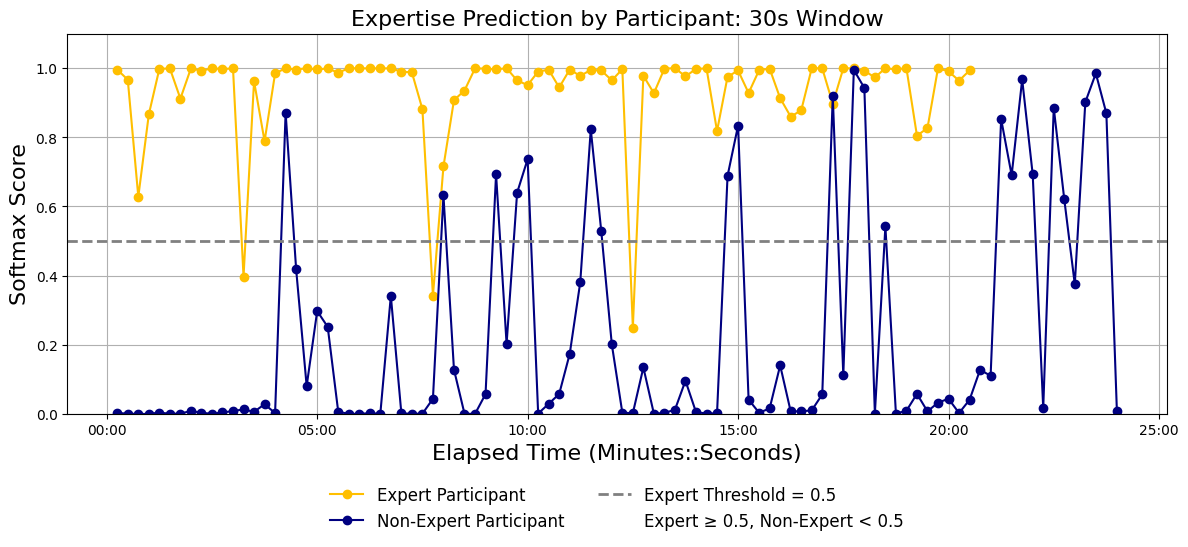}
    \vspace{1.0em}
    \includegraphics[width=0.9\linewidth]{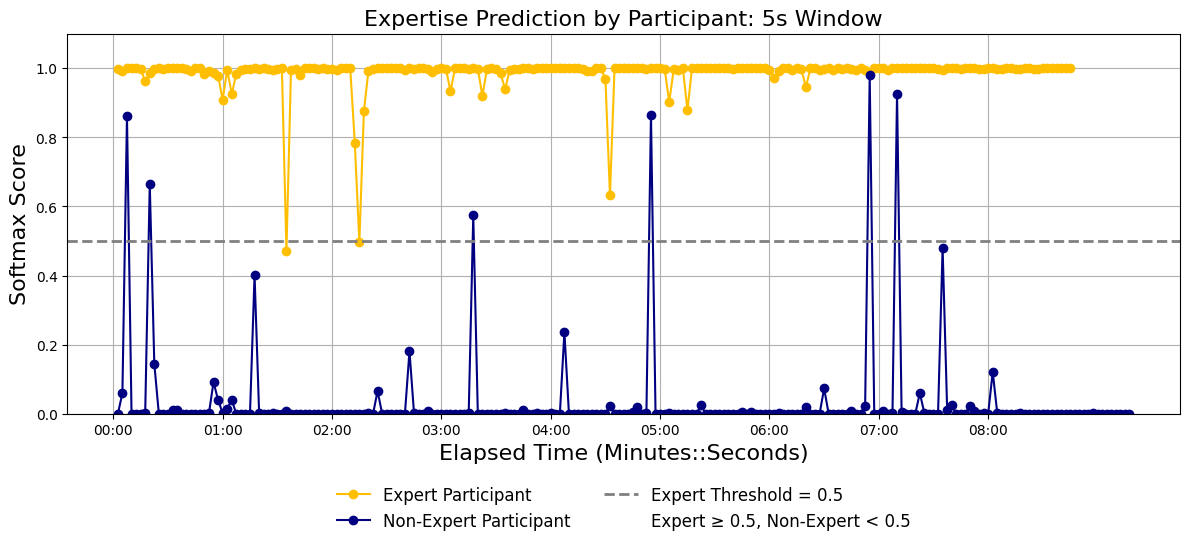}
    \caption{The predicted softmax scores of sequence data for both an expert and non-expert participant. {\bf Upper figure:} a test trial with a sliding window size of 30 seconds. {\bf Bottom figure:} a trial with a sliding window size of 5 seconds.}
    \label{fig:softmax_scores}
\end{figure*}

We trained the classifier for different window sizes (cf. Sec. \ref{sec:sliding_window}) and evaluated the results in batches of 12 models per window size to report both average results as well as the standard deviations from the mean error rates. 

\begin{table}[h!]
\centering
\caption{Mean AUROC scores along with their standard deviations \\for varying window sizes, obtained for windows sampled from both initial phase and verbal description phase.}
\label{tab:initial}
\begin{tabular}{c|c}
\toprule
\textbf{Window size (s)} & \textbf{Mean AUROC $\pm$ standard deviation} \\
\midrule
5  & 0.694 $\pm$ 0.237 \\
10 & 0.757 $\pm$ 0.212 \\
15 & 0.734 $\pm$ 0.226 \\
20 & 0.798 $\pm$ 0.219 \\
{\bf 30} & {\bf 0.831 $\pm$ 0.180} \\
\bottomrule
\end{tabular}
\end{table}

Table \ref{tab:initial} shows the mean AUROC scores along with their standard deviations for varying window sizes, obtained for windows sampled from both initial phase and verbal description phase. As it can be seen in the table, the mean AUROC score for the 5-second window size was initially \textbf{0.694 $\pm$ 0.237}. However, after analyzing the softmax scores of individual sequences for these results, we found statistically significant differences in the softmax scores of the expert class on sequences belonging to the initial evaluation phase vs. sequences belonging to the verbal description phase. Meaning that for small window sizes, sequences that correlated to the verbal description phase were not good predictors of the expertise. 

We thus retrained the models in the same format as described in the classifier design section, but limited the data to only sequences that belonged to the initial decision phase of the experiment. These models saw an increase in mean AUROC performance of \textbf{0.751 $\pm$ 0.237} for a 5-second window size. We noticed a similar trend in the data for a 10 second window size and retraining with only initial phase sequences saw an increase in mean AUROC performance to \textbf{0.780 $\pm$ 0.194}. We did not perform this analysis for the additional window sizes because their window size far exceeded the average time it took participants to give an initial decision, 9.5 seconds, and therefore there would not be enough sequence data for these window sizes to adequately train a model. Figure \ref{fig:ROC_Curves} shows the actual mean ROC curves with the corresponding AUROC scores for the final, best performing models for each window size. {\bf Thus, the answer to RQ2 is affirmative: we can train a classifier to automatically and in real-time distinguish experts and non-experts using eye tracking data from visual inspection tasks}.

\subsection{Answering \textbf{RQ3}: How long does a participant need to be observed by the system for an accurate expertise rating?}
According to the results of \textbf{RQ2}, we recommend that if an evaluation window of 30 seconds can be afforded (which translates to generating the classification result every 15 seconds), this sliding window length results in the best overall AUROC performance and offers the best expert evaluation overall. If a shorter sliding window length is needed for a mission-critical application (that is, shorter response time), there is not a significant decrease in the AUROC performance for the 5-, 10-, 15-, and 20-second window sizes.

\subsection{Performance of Experts vs. Non-Experts Recognition}

In this section, we report on the performance differences between experts and non-experts in the Iris PAD recognition task. Of the 54 image samples, experts had a correct mean of 46, a max of 52, and min of 39. Non-experts had a correct mean of 38, a max of 45, and a min of 32. These differences become more pronounced when performing a specific attack type analysis. There are 11 total categories of attack types in each deck given to a participant and the cumulative statistics from experts and non-experts show that 4 of these categories (artificial, contact lens samples printed, printouts, and post-mortem) boasted accuracies of over 90 percent. Five additional attack types showed success rates over 75 percent leaving the remaining three attack types (StyleGAN2-generated, bonafide irises, and StyleGAN3-generated) as the most difficult ones to identify with success rates of 48 percent, 55 percent, and 74 percent, respectively. Witnessing a high success rate on a number of categories, it is likely that some non-expert participants should be classified as experts from time to time, which is an added challenge. This point is illustrated in \cref{fig:softmax_scores}, which shows examples of the typical test results of the best performing configurations for a 5-second and 30-second window sizes. The figures show that on occasion, the non-expert participant will show expert-level performance in brief sequences or series of sequences.

\section{Discussion and Future Work}

Using common eye tracking metrics and gaze data from sequences of data in a sliding window approach, we built a classifier that distinguishes expert and non-expert performers. These eye tracking metrics have historically been used to differentiate between the performance of experts and non-experts while the gaze data indicates differences in domain-familiar visual search from exploratory search behavior in a variety of application areas including source code review, educational instruction, and a host of medical specialties \cite{hauser2023visual,van2014first,bernal2014experts,capogna2020novice,krupinski2013characterizing,li2023using}. Our results using the AutoSIGHT methodology did not consider eye tracking features calculated using AOIs related to our domain of choice. Instead we only used metrics and gaze patterns that have commonly been demonstrated to differentiate the expert and non-expert groups across a variety of domains. Therefore we can conclude that the results should generalize to a variety of domains.

We also found that certain expert/non-expert splits led to lower classification accuracy on data acquired from unseen subjects. This is a result of (1) some non-expert participants performing at expert levels for long periods of time, and (2) the primary separation of experts and non-experts was more apparent on certain attack types. Each set of images shown to the subjects contained easy-to-classify samples that allowed non-expert participants to potentially show high amounts of visual expertise on these images. That is, a non-expert participant may not be able to tell about the significant security vulnerabilities that irises wearing textured contacts create for iris recognition systems, but they can easily and expertly recognize the abnormal features of these samples. There are samples used during the model training where starting from an initial label of non-expert, the participant likely showed expert performance, and this adds an additional layer of complexity to separating the two groups. 

For improved results, {\bf future studies} should follow our blueprint in domains with less ambiguous expertise designations, for example, expert radiologists with many years of practice vs. radiologists in residency, or a more difficult biometric domain such as fingerprint PAD detection. This is especially true for domains that rely on, and frequently use, AOIs in their eye tracking studies. Seeing the positive performance of AutoSIGHT without the AOI metrics, it would be an interesting next step to see the baseline methodology of our approach fine-tuned to specific domains. The architecture used in the AutoSIGHT methodology can readily accommodate the addition of more eye tracking metrics, and by adding in AOI metrics such as dwell time, time to first fixation, revisits, etc., AutoSIGHT opens up pathways for future work in expert-guided human saliency experiments, continuous learning feedback loop experiments, and many others where it is beneficial to have expert-designated input and where expertise can now be demonstrated and verified. In the domain of code review, the ability to discern expert and non-expert behavior has the potential offer valuable insights into sections of code that may need extra attention, particularly if high percentages of non-expert interactions occur. We envision that AutoSIGHT could enable organizations to adopt a code review policy where a certain amount of expert interactions are required during the review before a merge request can be completed, thereby identifying potential problematic sections of code and preventing project disruption. As eye tracking becomes scalable to lower-end hardware such as webcams, we envision that AutoSIGHT's ability to monitor real-time assessment of expertise will enable dynamic content management in the education space. For instance, Morgan et al \cite{morgan2024investigating} identified ``preparation for complexity'' ({\it i.e.}, supporting students before frustration leads to disengagement) as a critical factor for coding education for children. AutoSIGHT has the potential to integrate into their future work ideas by using eye tracking to continuously and in real-time monitor the performance of students and alert the instructor to struggling students enabling timely interventions. Additionally, AutoSIGHT can be used to dynamically adjust the workload if a student continually scores too low or too high.

Finally, AutoSIGHT offers the potential to open up new pathways for better Human-Centric AI tools. The eye tracking heatmaps obtained from participants, that are released along with this paper, can be used for saliency-guided model training paradigms such as \cite{Boyd_2023_WACV}. AutoSIGHT can automatically assess the visual expertise and thus allow for selection of human saliency maps which are delivered by the most qualified experts. This may lead to much better saliency data used in the saliency-guided model training, increasing the fidelity of explanations and generalization of AI tools supporting human experts.

\section{Conclusions}
This paper proposes AutoSIGHT, a methodology and classifier architecture to assess visual expertise in real time allowing for an expertise-weighted AI-human teaming, in which the dynamics of the expertise level of both players (AI or human) may change rapidly (\eg due to the need of replacing a human expert). In our study, features consisting of gaze and eye tracking data were segmented into various sliding window sizes to test how long a participant needed to be observed before an accurate expertise prediction could be made. This was accomplished via a multi-stream neural network that achieved mean AUROC of 0.751 in classifying visual expertise for a sequence length of just 5 seconds in a subject-disjoint train-test scenario. Furthermore, when evaluating window sizes of 30 seconds, the AUROC performance increases to 0.831. Finally, we discussed some shortcomings related to our domain of choice (iris presentation attack detection) and discussed ways for future researchers to adapt this work to other domains. To facilitate the application of this work and follow-up research, the collected dataset, source codes, and model weights are offered along with this paper.

\section*{Acknowledgment} This work was supported by the U.S. Department of Defense (Contract No. W52P1J-20-9-3009). Any opinions, findings, and conclusions or recommendations expressed in this material are those of the authors and do not necessarily reflect the views of the U.S. Department of Defense or the U.S. Government. The U.S. Government is authorized to reproduce and distribute reprints for Government purposes, notwithstanding any copyright notation here on.

\bibliographystyle{IEEEtran}

\begin{thebibliography}{10}
\providecommand{\url}[1]{#1}
\csname url@samestyle\endcsname
\providecommand{\newblock}{\relax}
\providecommand{\bibinfo}[2]{#2}
\providecommand{\BIBentrySTDinterwordspacing}{\spaceskip=0pt\relax}
\providecommand{\BIBentryALTinterwordstretchfactor}{4}
\providecommand{\BIBentryALTinterwordspacing}{\spaceskip=\fontdimen2\font plus
\BIBentryALTinterwordstretchfactor\fontdimen3\font minus \fontdimen4\font\relax}
\providecommand{\BIBforeignlanguage}[2]{{%
\expandafter\ifx\csname l@#1\endcsname\relax
\typeout{** WARNING: IEEEtran.bst: No hyphenation pattern has been}%
\typeout{** loaded for the language `#1'. Using the pattern for}%
\typeout{** the default language instead.}%
\else
\language=\csname l@#1\endcsname
\fi
#2}}
\providecommand{\BIBdecl}{\relax}
\BIBdecl

\bibitem{hasija2022artificial}
A.~Hasija and T.~L. Esper, ``In artificial intelligence (ai) we trust: A qualitative investigation of ai technology acceptance,'' \emph{Journal of Business Logistics}, vol.~43, no.~3, pp. 388--412, 2022.

\bibitem{Dang_2020_CVPR}
H.~Dang, F.~Liu, J.~Stehouwer, X.~Liu, and A.~K. Jain, ``On the detection of digital face manipulation,'' in \emph{Proceedings of the IEEE/CVF Conference on Computer Vision and Pattern Recognition (CVPR)}, June 2020.

\bibitem{Boyd_2023_WACV}
A.~Boyd, P.~Tinsley, K.~W. Bowyer, and A.~Czajka, ``Cyborg: Blending human saliency into the loss improves deep learning-based synthetic face detection,'' in \emph{Proceedings of the IEEE/CVF Winter Conference on Applications of Computer Vision (WACV)}, January 2023, pp. 6108--6117.

\bibitem{crum2023explain}
C.~R. Crum, P.~Tinsley, A.~Boyd, J.~Piland, C.~Sweet, T.~Kelley, K.~Bowyer, and A.~Czajka, ``Explain to me: Salience-based explainability for synthetic face detection models,'' \emph{IEEE Transactions on Artificial Intelligence}, 2023.

\bibitem{boyd2022human}
A.~Boyd, K.~W. Bowyer, and A.~Czajka, ``Human-aided saliency maps improve generalization of deep learning,'' in \emph{Proceedings of the IEEE/CVF Winter Conference on Applications of Computer Vision}, 2022, pp. 2735--2744.

\bibitem{Sultana_2024_ACCV}
J.~Sultana, R.~Qin, and Z.~Yin, ``Seeing through expert's eyes: Leveraging radiologist eye gaze and speech report with graph neural networks for chest x-ray image classification,'' in \emph{Proceedings of the Asian Conference on Computer Vision (ACCV)}, December 2024, pp. 2579--2595.

\bibitem{kaushal2023detecting}
S.~Kaushal, Y.~Sun, R.~Zukerman, R.~W. Chen, and K.~A. Thakoor, ``Detecting eye disease using vision transformers informed by ophthalmology resident gaze data,'' in \emph{2023 45th Annual International Conference of the IEEE Engineering in Medicine \& Biology Society (EMBC)}.\hskip 1em plus 0.5em minus 0.4em\relax IEEE, 2023, pp. 1--4.

\bibitem{samei2018handbook}
E.~Samei and E.~A. Krupinski, \emph{The handbook of medical image perception and techniques}.\hskip 1em plus 0.5em minus 0.4em\relax Cambridge University Press, 2018.

\bibitem{krupinski2013characterizing}
E.~A. Krupinski, A.~R. Graham, and R.~S. Weinstein, ``Characterizing the development of visual search expertise in pathology residents viewing whole slide images,'' \emph{Human pathology}, vol.~44, no.~3, pp. 357--364, 2013.

\bibitem{kelly2016development}
B.~S. Kelly, L.~A. Rainford, S.~P. Darcy, E.~C. Kavanagh, and R.~J. Toomey, ``The development of expertise in radiology: in chest radiograph interpretation,“expert” search pattern may predate “expert” levels of diagnostic accuracy for pneumothorax identification,'' \emph{Radiology}, vol. 280, no.~1, pp. 252--260, 2016.

\bibitem{hobbes2016leviathan}
T.~Hobbes, ``Leviathan,'' in \emph{Democracy: a reader}.\hskip 1em plus 0.5em minus 0.4em\relax Columbia University Press, 2016, pp. 37--42.

\bibitem{thiebes2021trustworthy}
S.~Thiebes, S.~Lins, and A.~Sunyaev, ``Trustworthy artificial intelligence,'' \emph{Electronic Markets}, vol.~31, pp. 447--464, 2021.

\bibitem{smuha2019eu}
N.~A. Smuha, ``The eu approach to ethics guidelines for trustworthy artificial intelligence,'' \emph{Computer Law Review International}, vol.~20, no.~4, pp. 97--106, 2019.

\bibitem{brundage2020trustworthy}
M.~Brundage, S.~Avin, J.~Wang, H.~Belfield, G.~Krueger, G.~Hadfield, H.~Khlaaf, J.~Yang, H.~Toner, R.~Fong, T.~Maharaj, P.~W. Koh, S.~Hooker, J.~Leung, A.~Trask, E.~Bluemke, J.~Lebensold, C.~O'Keefe, M.~Koren, T.~Ryffel, J.~Rubinovitz, T.~Besiroglu, F.~Carugati, J.~Clark, P.~Eckersley, S.~de~Haas, M.~Johnson, B.~Laurie, A.~Ingerman, I.~Krawczuk, A.~Askell, R.~Cammarota, A.~Lohn, D.~Krueger, C.~Stix, P.~Henderson, L.~Graham, C.~Prunkl, B.~Martin, E.~Seger, N.~Zilberman, S.~O. Eigeartaigh, F.~Kroeger, G.~Sastry, R.~Kagan, A.~Weller, B.~Tse, E.~Barnes, A.~Dafoe, P.~Scharre, A.~Herbert-Voss, M.~Rasser, S.~Sodhani, C.~Flynn, T.~K. Gilbert, L.~Dyer, S.~Khan, Y.~Bengio, and M.~Anderljung, ``Toward trustworthy ai development: Mechanisms for supporting verifiable claims,'' 2020.

\bibitem{10.1145/3491209}
\BIBentryALTinterwordspacing
D.~Kaur, S.~Uslu, K.~J. Rittichier, and A.~Durresi, ``Trustworthy artificial intelligence: A review,'' \emph{ACM Comput. Surv.}, vol.~55, no.~2, jan 2022. [Online]. Available: \url{https://doi.org/10.1145/3491209}
\BIBentrySTDinterwordspacing

\bibitem{kaur2021requirements}
D.~Kaur, S.~Uslu, and A.~Durresi, ``Requirements for trustworthy artificial intelligence--a review,'' in \emph{Advances in Networked-Based Information Systems: The 23rd International Conference on Network-Based Information Systems (NBiS-2020) 23}.\hskip 1em plus 0.5em minus 0.4em\relax Springer, 2021, pp. 105--115.

\bibitem{li2023trustworthy}
B.~Li, P.~Qi, B.~Liu, S.~Di, J.~Liu, J.~Pei, J.~Yi, and B.~Zhou, ``Trustworthy ai: From principles to practices,'' \emph{ACM Computing Surveys}, vol.~55, no.~9, pp. 1--46, 2023.

\bibitem{liu2022trustworthy}
H.~Liu, Y.~Wang, W.~Fan, X.~Liu, Y.~Li, S.~Jain, Y.~Liu, A.~Jain, and J.~Tang, ``Trustworthy ai: A computational perspective,'' \emph{ACM Transactions on Intelligent Systems and Technology}, vol.~14, no.~1, pp. 1--59, 2022.

\bibitem{ribeiro2016should}
M.~T. Ribeiro, S.~Singh, and C.~Guestrin, ``" why should i trust you?" explaining the predictions of any classifier,'' in \emph{Proceedings of the 22nd ACM SIGKDD international conference on knowledge discovery and data mining}, 2016, pp. 1135--1144.

\bibitem{castelo2019task}
N.~Castelo, M.~W. Bos, and D.~R. Lehmann, ``Task-dependent algorithm aversion,'' \emph{Journal of Marketing Research}, vol.~56, no.~5, pp. 809--825, 2019.

\bibitem{longoni2019resistance}
C.~Longoni, A.~Bonezzi, and C.~K. Morewedge, ``Resistance to medical artificial intelligence,'' \emph{Journal of Consumer Research}, vol.~46, no.~4, pp. 629--650, 2019.

\bibitem{longoni2020resistance}
------, ``Resistance to medical artificial intelligence is an attribute in a compensatory decision process: response to pezzo and beckstead (2020),'' \emph{Judgment and Decision Making}, vol.~15, no.~3, pp. 446--448, 2020.

\bibitem{robertson2023diverse}
C.~Robertson, A.~Woods, K.~Bergstrand, J.~Findley, C.~Balser, and M.~J. Slepian, ``Diverse patients’ attitudes towards artificial intelligence (ai) in diagnosis,'' \emph{PLOS Digital Health}, vol.~2, no.~5, p. e0000237, 2023.

\bibitem{riedl2024patients}
R.~Riedl, S.~A. Hogeterp, and M.~Reuter, ``Do patients prefer a human doctor, artificial intelligence, or a blend, and is this preference dependent on medical discipline? empirical evidence and implications for medical practice,'' \emph{Frontiers in Psychology}, vol.~15, p. 1422177, 2024.

\bibitem{jiang2018trust}
H.~Jiang, B.~Kim, M.~Guan, and M.~Gupta, ``To trust or not to trust a classifier,'' \emph{Advances in neural information processing systems}, vol.~31, 2018.

\bibitem{nguyen2015deep}
A.~Nguyen, J.~Yosinski, and J.~Clune, ``Deep neural networks are easily fooled: High confidence predictions for unrecognizable images,'' in \emph{Proceedings of the IEEE conference on computer vision and pattern recognition}, 2015, pp. 427--436.

\bibitem{selvaraju2017grad}
R.~R. Selvaraju, M.~Cogswell, A.~Das, R.~Vedantam, D.~Parikh, and D.~Batra, ``Grad-cam: Visual explanations from deep networks via gradient-based localization,'' in \emph{Proceedings of the IEEE international conference on computer vision}, 2017, pp. 618--626.

\bibitem{draelos2021use}
R.~L. Draelos and L.~Carin, ``Use hirescam instead of grad-cam for faithful explanations of convolutional neural networks,'' 2021.

\bibitem{saporta2022benchmarking}
A.~Saporta, X.~Gui, A.~Agrawal, A.~Pareek, S.~Q. Truong, C.~D. Nguyen, V.-D. Ngo, J.~Seekins, F.~G. Blankenberg, A.~Y. Ng \emph{et~al.}, ``Benchmarking saliency methods for chest x-ray interpretation,'' \emph{Nature Machine Intelligence}, vol.~4, no.~10, pp. 867--878, 2022.

\bibitem{boyd2023value}
A.~Boyd, P.~Tinsley, K.~Bowyer, and A.~Czajka, ``The value of ai guidance in human examination of synthetically-generated faces,'' in \emph{Proceedings of the AAAI Conference on Artificial Intelligence}, vol.~37, no.~5, 2023, pp. 5930--5938.

\bibitem{bourne2014expertise}
L.~E. Bourne~Jr, J.~A. Kole, and A.~F. Healy, ``Expertise: defined, described, explained,'' \emph{Frontiers in psychology}, vol.~5, p. 186, 2014.

\bibitem{jarodzka2021eye}
H.~Jarodzka, I.~Skuballa, and H.~Gruber, ``Eye-tracking in educational practice: Investigating visual perception underlying teaching and learning in the classroom,'' \emph{Educational Psychology Review}, vol.~33, no.~1, pp. 1--10, 2021.

\bibitem{borys2017eye}
M.~Borys and M.~Plechawska-W{\'o}jcik, ``Eye-tracking metrics in perception and visual attention research,'' \emph{EJMT}, vol.~3, pp. 11--23, 2017.

\bibitem{tien2014eye}
T.~Tien, P.~H. Pucher, M.~H. Sodergren, K.~Sriskandarajah, G.-Z. Yang, and A.~Darzi, ``Eye tracking for skills assessment and training: a systematic review,'' \emph{journal of surgical research}, vol. 191, no.~1, pp. 169--178, 2014.

\bibitem{gegenfurtner2011expertise}
A.~Gegenfurtner, E.~Lehtinen, and R.~S{\"a}lj{\"o}, ``Expertise differences in the comprehension of visualizations: A meta-analysis of eye-tracking research in professional domains,'' \emph{Educational psychology review}, vol.~23, pp. 523--552, 2011.

\bibitem{kosel2023measuring}
C.~Kosel, A.~Mooseder, T.~Seidl, and J.~Pfeffer, ``Measuring teachers' visual expertise using the gaze relational index based on real-world eye-tracking data and varying velocity thresholds,'' \emph{arXiv preprint arXiv:2304.05143}, 2023.

\bibitem{castner2020pupil}
N.~Castner, T.~Appel, T.~Eder, J.~Richter, K.~Scheiter, C.~Keutel, F.~H{\"u}ttig, A.~Duchowski, and E.~Kasneci, ``Pupil diameter differentiates expertise in dental radiography visual search,'' \emph{PloS one}, vol.~15, no.~5, p. e0223941, 2020.

\bibitem{bednarik2018pupil}
R.~Bednarik, P.~Bartczak, H.~Vrzakova, J.~Koskinen, A.-P. Elomaa, A.~Huotarinen, D.~G. de~G{\'o}mez~P{\'e}rez, and M.~von und~zu Fraunberg, ``Pupil size as an indicator of visual-motor workload and expertise in microsurgical training tasks,'' in \emph{Proceedings of the 2018 ACM Symposium on Eye Tracking Research \& Applications}, 2018, pp. 1--5.

\bibitem{li2023using}
S.~Li, M.~C. Duffy, S.~P. Lajoie, J.~Zheng, and K.~Lachapelle, ``Using eye tracking to examine expert-novice differences during simulated surgical training: A case study,'' \emph{Computers in Human Behavior}, vol. 144, p. 107720, 2023.

\bibitem{brunye2019eye}
T.~T. Bruny{\'e}, B.~K. Nallamothu, and J.~G. Elmore, ``Eye-tracking for assessing medical image interpretation: A pilot feasibility study comparing novice vs expert cardiologists,'' \emph{Perspectives on medical education}, vol.~8, pp. 65--73, 2019.

\bibitem{bernal2014experts}
J.~Bernal, F.~J. S{\'a}nchez, F.~Vilarino, M.~Arnold, A.~Ghosh, and G.~Lacey, ``Experts vs. novices: applying eye-tracking methodologies in colonoscopy video screening for polyp search,'' in \emph{Proceedings of the symposium on eye tracking research and applications}, 2014, pp. 223--226.

\bibitem{capogna2020novice}
E.~Capogna, F.~Salvi, L.~Delvino, A.~Di~Giacinto, and M.~Velardo, ``Novice and expert anesthesiologists’ eye-tracking metrics during simulated epidural block: a preliminary, brief observational report,'' \emph{Local and Regional Anesthesia}, pp. 105--109, 2020.

\bibitem{ooms2014study}
K.~Ooms, P.~De~Maeyer, and V.~Fack, ``Study of the attentive behavior of novice and expert map users using eye tracking,'' \emph{Cartography and Geographic Information Science}, vol.~41, no.~1, pp. 37--54, 2014.

\bibitem{dogusoy2014cognitive}
B.~Dogusoy-Taylan and K.~Cagiltay, ``Cognitive analysis of experts’ and novices’ concept mapping processes: An eye tracking study,'' \emph{Computers in human behavior}, vol.~36, pp. 82--93, 2014.

\bibitem{boyd2020iris}
A.~Boyd, Z.~Fang, A.~Czajka, and K.~W. Bowyer, ``Iris presentation attack detection: Where are we now?'' \emph{Pattern Recognition Letters}, vol. 138, pp. 483--489, 2020.

\bibitem{czajka2018presentation}
A.~Czajka and K.~W. Bowyer, ``Presentation attack detection for iris recognition: An assessment of the state-of-the-art,'' \emph{ACM Computing Surveys (CSUR)}, vol.~51, no.~4, pp. 1--35, 2018.

\bibitem{daugman2009iris}
J.~Daugman, ``How iris recognition works,'' in \emph{The essential guide to image processing}.\hskip 1em plus 0.5em minus 0.4em\relax Elsevier, 2009, pp. 715--739.

\bibitem{boyd2023comprehensive}
A.~Boyd, J.~Speth, L.~Parzianello, K.~W. Bowyer, and A.~Czajka, ``Comprehensive study in open-set iris presentation attack detection,'' \emph{IEEE Transactions on Information Forensics and Security}, vol.~18, pp. 3238--3250, 2023.

\bibitem{pacut2006aliveness}
A.~Pacut and A.~Czajka, ``Aliveness detection for iris biometrics,'' in \emph{Proceedings 40th annual 2006 international carnahan conference on security technology}.\hskip 1em plus 0.5em minus 0.4em\relax IEEE, 2006, pp. 122--129.

\bibitem{trokielewicz2018presentation}
M.~Trokielewicz, A.~Czajka, and P.~Maciejewicz, ``Presentation attack detection for cadaver iris,'' in \emph{2018 IEEE 9th international conference on biometrics theory, applications and systems (BTAS)}.\hskip 1em plus 0.5em minus 0.4em\relax IEEE, 2018, pp. 1--10.

\bibitem{Yadav_2019_CVPR_Workshops}
S.~Yadav, C.~Chen, and A.~Ross, ``Synthesizing iris images using rasgan with application in presentation attack detection,'' in \emph{Proceedings of the IEEE/CVF Conference on Computer Vision and Pattern Recognition (CVPR) Workshops}, June 2019.

\bibitem{karras2020analyzing}
T.~Karras, S.~Laine, M.~Aittala, J.~Hellsten, J.~Lehtinen, and T.~Aila, ``Analyzing and improving the image quality of stylegan,'' in \emph{Proceedings of the IEEE/CVF conference on computer vision and pattern recognition}, 2020, pp. 8110--8119.

\bibitem{karras2021alias}
T.~Karras, M.~Aittala, S.~Laine, E.~H{\"a}rk{\"o}nen, J.~Hellsten, J.~Lehtinen, and T.~Aila, ``Alias-free generative adversarial networks,'' \emph{Advances in neural information processing systems}, vol.~34, pp. 852--863, 2021.

\bibitem{menotti2015deep}
D.~Menotti, G.~Chiachia, A.~Pinto, W.~R. Schwartz, H.~Pedrini, A.~X. Falcao, and A.~Rocha, ``Deep representations for iris, face, and fingerprint spoofing detection,'' \emph{IEEE Transactions on Information Forensics and Security}, vol.~10, no.~4, pp. 864--879, 2015.

\bibitem{minaee2016experimental}
S.~Minaee, A.~Abdolrashidiy, and Y.~Wang, ``An experimental study of deep convolutional features for iris recognition,'' in \emph{2016 IEEE signal processing in medicine and biology symposium (SPMB)}.\hskip 1em plus 0.5em minus 0.4em\relax IEEE, 2016, pp. 1--6.

\bibitem{alaslani2018convolutional}
M.~G. Alaslani, ``Convolutional neural network based feature extraction for iris recognition,'' \emph{International Journal of Computer Science \& Information Technology (IJCSIT) Vol}, vol.~10, 2018.

\bibitem{boyd2019deep}
A.~Boyd, A.~Czajka, and K.~Bowyer, ``Deep learning-based feature extraction in iris recognition: Use existing models, fine-tune or train from scratch?'' in \emph{2019 IEEE 10th International Conference on Biometrics Theory, Applications and Systems (BTAS)}.\hskip 1em plus 0.5em minus 0.4em\relax IEEE, 2019, pp. 1--9.

\bibitem{kim2019saliency}
B.~Kim, J.~Seo, S.~Jeon, J.~Koo, J.~Choe, and T.~Jeon, ``Why are saliency maps noisy? cause of and solution to noisy saliency maps,'' in \emph{2019 IEEE/CVF International Conference on Computer Vision Workshop (ICCVW)}.\hskip 1em plus 0.5em minus 0.4em\relax IEEE, 2019, pp. 4149--4157.

\bibitem{Kohli_BTAS_2016}
N.~Kohli, D.~Yadav, M.~Vatsa, R.~Singh, and A.~Noore, ``Detecting medley of iris spoofing attacks using desist,'' in \emph{2016 IEEE 8th International Conference on Biometrics Theory, Applications and Systems (BTAS)}, 2016, pp. 1--6.

\bibitem{CASIA_Synth}
\BIBentryALTinterwordspacing
T.~Tan, ``Casia-iris-syn (part of casia-irisv4),'' 2024. [Online]. Available: \url{https://hycasia.github.io/dataset/casia-irisv4/}
\BIBentrySTDinterwordspacing

\bibitem{Trokielewicz_BTAS_2015}
M.~Trokielewicz, A.~Czajka, and P.~Maciejewicz, ``Database of iris images acquired in the presence of ocular pathologies and assessment of iris recognition reliability for disease-affected eyes,'' in \emph{2015 IEEE 2nd International Conference on Cybernetics (CYBCONF)}, 2015, pp. 495--500.

\bibitem{Trokielewicz_TIFS_2019}
------, ``Iris recognition after death,'' \emph{IEEE Transactions on Information Forensics and Security}, vol.~14, no.~6, pp. 1501--1514, 2019.

\bibitem{Czajka_MMAR_2013}
A.~Czajka, ``Database of iris printouts and its application: Development of liveness detection method for iris recognition,'' in \emph{2013 18th International Conference on Methods and Models in Automation and Robotics (MMAR)}, 2013, pp. 28--33.

\bibitem{Doyle_IPBR_2014}
J.~Doyle and K.~Bowyer, ``Notre dame image dataset for contact lens detection in iris recognition,'' \emph{Iris and periocular biometric recognition}, pp. 265--290, 2014.

\bibitem{Kim_ESA_2016}
\BIBentryALTinterwordspacing
D.~Kim, Y.~Jung, K.-A. Toh, B.~Son, and J.~Kim, ``An empirical study on iris recognition in a mobile phone,'' \emph{Expert Systems with Applications}, vol.~54, pp. 328--339, 2016. [Online]. Available: \url{https://www.sciencedirect.com/science/article/pii/S0957417416300148}
\BIBentrySTDinterwordspacing

\bibitem{dunn2024minimal}
M.~J. Dunn, R.~G. Alexander, O.~M. Amiebenomo, G.~Arblaster, D.~Atan, J.~T. Erichsen, U.~Ettinger, M.~E. Giardini, I.~D. Gilchrist, R.~Hamilton \emph{et~al.}, ``Minimal reporting guideline for research involving eye tracking (2023 edition),'' \emph{Behavior research methods}, vol.~56, no.~5, pp. 4351--4357, 2024.

\bibitem{tinsley2023iris}
P.~Tinsley, S.~Purnapatra, M.~Mitcheff, A.~Boyd, C.~Crum, K.~Bowyer, P.~Flynn, S.~Schuckers, A.~Czajka, M.~Fang \emph{et~al.}, ``Iris liveness detection competition (livdet-iris)--the 2023 edition,'' in \emph{2023 IEEE International Joint Conference on Biometrics (IJCB)}.\hskip 1em plus 0.5em minus 0.4em\relax IEEE, 2023, pp. 1--10.

\bibitem{gegenfurtner2020gaze}
A.~Gegenfurtner, J.-M. Boucheix, H.~Gruber, F.~Hauser, E.~Lehtinen, and R.~K. Lowe, ``The gaze relational index as a measure of visual expertise,'' 2020.

\bibitem{szulewski2017pupillometry}
A.~Szulewski, D.~Kelton, and D.~Howes, ``Pupillometry as a tool to study expertise in medicine.'' \emph{Frontline Learning Research}, vol.~5, no.~3, pp. 53--63, 2017.

\bibitem{ruuskanen2023baseline}
V.~Ruuskanen, T.~Hagen, T.~Espeseth, and S.~Math{\^o}t, ``Baseline pupil size seems unrelated to fluid intelligence, working memory capacity, and attentional control,'' \emph{bioRxiv}, pp. 2023--06, 2023.

\bibitem{He_CVPR_2016}
K.~He, X.~Zhang, S.~Ren, and J.~Sun, ``Deep residual learning for image recognition,'' in \emph{Proceedings of the IEEE Conference on Computer Vision and Pattern Recognition (CVPR)}, 2016, pp. 770--778.

\bibitem{rayner1998eye}
K.~Rayner, ``Eye movements in reading and information processing: 20 years of research.'' \emph{Psychological bulletin}, vol. 124, no.~3, p. 372, 1998.

\bibitem{negi2020fixation}
S.~Negi and R.~Mitra, ``Fixation duration and the learning process: An eye tracking study with subtitled videos,'' \emph{Journal of Eye Movement Research}, vol.~13, no.~6, 2020.

\bibitem{hauser2023visual}
F.~Hauser, L.~Grabinger, J.~Mottok, and H.~Gruber, ``Visual expertise in code reviews: using holistic models of image perception to analyze and interpret eye movements,'' in \emph{Proceedings of the 2023 Symposium on Eye Tracking Research and Applications}, 2023, pp. 1--7.

\bibitem{van2014first}
N.~van~den Bogert, J.~van Bruggen, D.~Kostons, and W.~Jochems, ``First steps into understanding teachers' visual perception of classroom events,'' \emph{Teaching and teacher education}, vol.~37, pp. 208--216, 2014.

\bibitem{morgan2024investigating}
M.~Morgan, K.~Gash, S.~Ludi, and T.~Truong, ``Investigating the usability of coding applications for children: Insights from teacher interviews,'' in \emph{2024 IEEE Symposium on Visual Languages and Human-Centric Computing (VL/HCC)}.\hskip 1em plus 0.5em minus 0.4em\relax IEEE, 2024, pp. 47--58.

\end{thebibliography}

\end{document}